\documentclass{article}

\usepackage[english]{babel}
\usepackage{float}
\usepackage{multirow}

\usepackage[letterpaper,top=2cm,bottom=2cm,left=3cm,right=3cm,marginparwidth=1.75cm]{geometry}

\usepackage{amsmath}
\usepackage{graphicx}
\usepackage[colorlinks=true, allcolors=blue]{hyperref}

\title{{Using Analytics on Student Created Data to Content Validate Pedagogical Tools}}

\begin{document}

\maketitle

\author{John Kos, jkos3@gatech.edu, Georgia Institute of Technology, Atlanta Georgia\\}
\author{Kenneth Eaton, keaton30@gatech.edu, Georgia Institute of Technology, Atlanta Georgia\\}
\author{Sareen Zhang, szhang702@gatech.edu, Georgia Institute of Technology, Atlanta Georgia\\}
\author{Rahul Dass, rdass7@gatech.edu, Georgia Institute of Technology, Atlanta Georgia\\}
\author{Stephen Buckley, sbuckley@gatech.edu, Georgia Institute of Technology, Atlanta Georgia\\}
\author{Sungeun An, sungeun.an@gatech.edu, Georgia Institute of Technology, Atlanta Georgia\\}
\author{Ashok Goel, ashok.goel@cc.gatech.edu, Georgia Institute of Technology, Atlanta Georgia\\}

\begin{abstract}
Conceptual and simulation models can function as useful pedagogical tools, however it is important to categorize different outcomes when evaluating them  in order to more meaningfully interpret results.                           
VERA is a ecology-based conceptual modeling software that enables users to simulate interactions between biotics and abiotics in an ecosystem, allowing users to form and then verify hypothesis through observing a time series of the species populations.
In this paper, we classify this time series into common patterns found in the domain of ecological modeling through two methods, hierarchical clustering and curve fitting, illustrating a general methodology for showing content validity when combining different pedagogical tools.
When applied to a diverse sample of 263 models containing 971 time series collected from three different VERA user categories: a Georgia Tech (GATECH), North Georgia Technical College (NGTC), and ``Self Directed Learners'', results showed agreement between both classification methods on 89.38\% of the sample curves in the test set. This serves as a good indication that our methodology for determining content validity was successful.
\end{abstract}

\section{Introduction}
Students in the studies of ecology and biology often need tools to make sense of the complex system found in real-world ecology. 
These pedagogical tools, often necessary for learning, simplify real-world phenomena and make it easier to learn and understand the myriad of factors affecting real-world circumstances can be difficult to capture with empirical approaches. Thus, modeling ecological systems may be a more tractable approach \cite{evans2012predictive}.

Among the tools used in ecological education are mathematical models, conceptual models, and simulations.
All three of these tools offer different insight to help students understand complex ecological phenomena.
Mathematical models are especially useful in describing individual population dynamics that delineate the growth or decline of species' population over  time, given various factors such as birth rate, death rate, etc.
This description of species' population over time often gets categorized into common curves in order to relate them to phenomena often found in nature \cite{eberhardt2008curves}.
However, when engaging in systems thinking, mathematical models become exponentially more complex and challenging to understand \cite{pielou1981stock}.

Conceptual models represent the behavior of a system by defining its components, relations, and processes \cite{an2018vera}. 
This allows students, through predictive inquiry, to formulate "what if" hypotheses that can be formalized with a conceptual model.
Although conceptual models provide an overview of a system and formulate a hypothesis, they cannot evaluate how those dynamics affect the system over time.
Simulation models provide a solution to this by using complex algorithms to model ecosystem dynamics and create the most accurate predictions \cite{2022Modeling}. 
In short, simulations provide the ability to evaluate predictive inquiry.

The Virtual Ecological Research Assistant (VERA) \cite{vera2023site} is an ecological modeling software that allows for an end-to-end evaluation process from defining the conceptual model, see Fig. 1, to observing the simulation outputs, see Fig. 2 \cite{an2018vera}.
This provides a near-instantaneous loop of hypothesis formation and evaluation.
As outputs, VERA displays changes in the species' populations over time. 
In Section \ref{sec2:a}, we describe how the VERA application works in more detail.

\begin{figure}[H]
    \centering
    \includegraphics[width=1\linewidth]{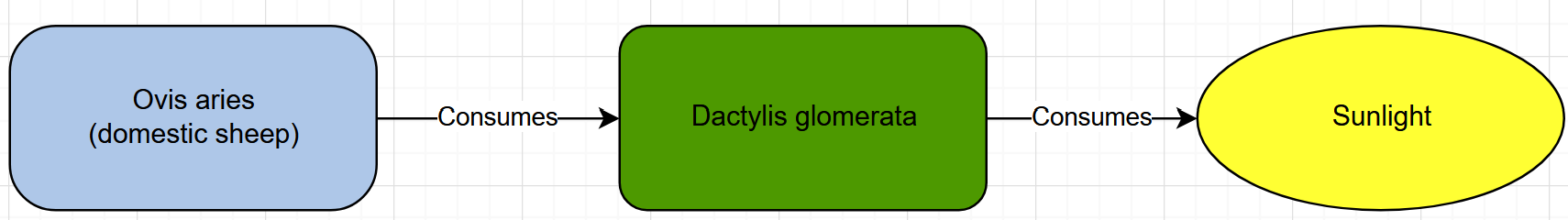}
    \caption{An example of a predator-prey conceptual model in VERA.}
    \label{fig:conceptual_model}
\end{figure}

\begin{figure}[H]
    \centering
    \includegraphics[width=.8\linewidth]{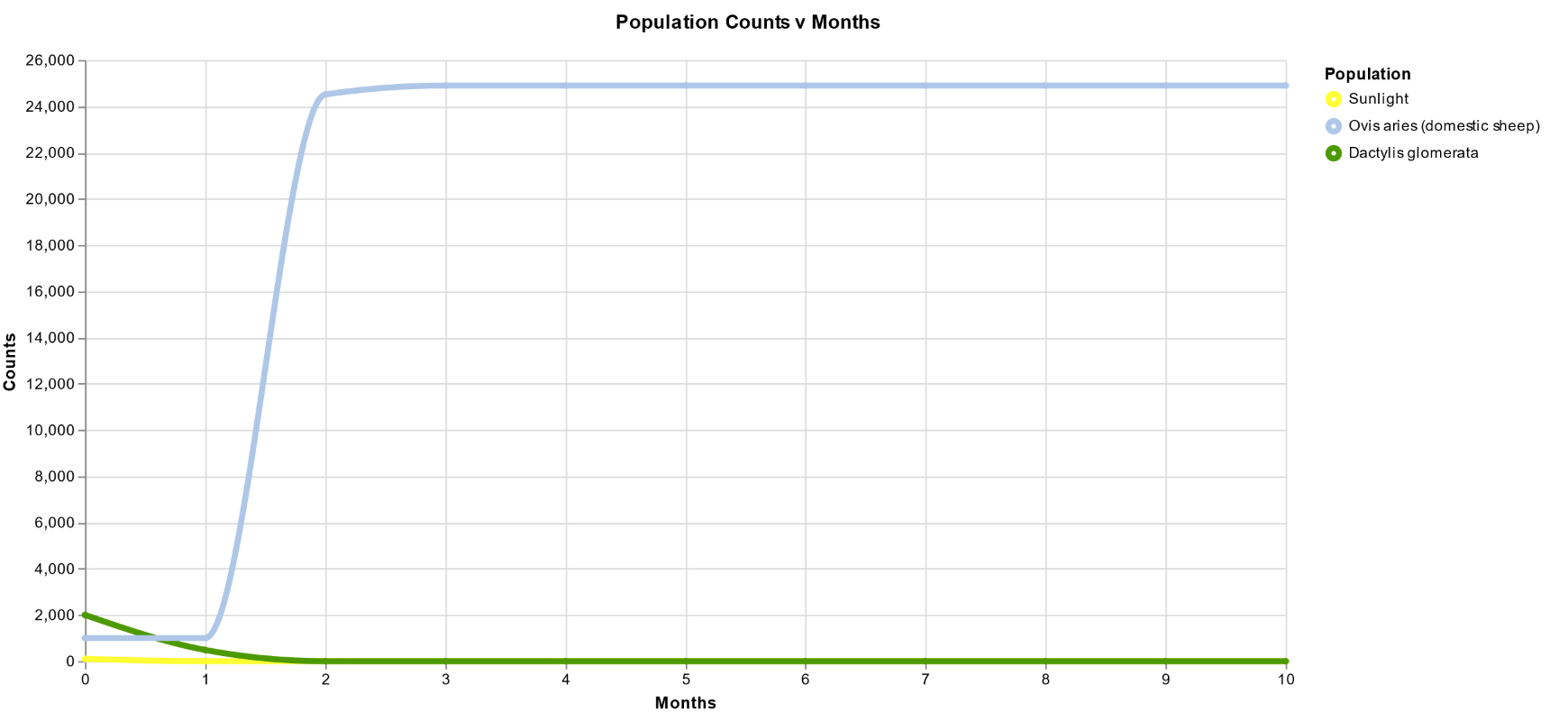}
    \caption{An example of the simulation output from a conceptual model in VERA}
    \label{fig:sub2}
\end{figure}

Simulations for pedagogical purposes are an increasingly popular method for teaching students across disciplines, including but not limited to social-ecology, epidemiology, and economics \cite{janssen2014simulation,dayblack2015simulation,delatorre2021simulation}.
Despite its popularity, there is a gap in educational simulation systems research with concern to assessing the internal content validity of the simulation based educational tools \cite{cook2016validity}.
By corroborating the expected behavior of encountering a known group of curves found in ecological modeling with the unsupervised bottom up approach of hierarchical clustering, we are proposing a methodology for validating the use of simulations used in education, with an emphasis primarily on domains with a heavy focus on modeling.

This methodology is also particularly useful for agent-based simulation systems for which the complexity of the model may cause outcomes unaligned with the educational purpose. 
This is due to the fact that the agent based system in VERA is attempting to simulate the complex system of ecology.
Complex systems are defined by Ottino \cite{ottino2003complex} as ``a system with a large number of elements, building blocks or agents, capable of exchanging stimuli with one another and with their environment''. 
Interactions within complex systems create emergent patterns.
Although VERA takes many fundamental principles from ecology, that does not necessarily guarantee that the emergent patterns will be the same. 
Ottino \cite{ottino2003complex} continues ``The common characteristic of all complex systems is that they display organization without any external organizing principle being applied.''
The convergence of expected outcomes between modeling--based disciplines and agent--based simulations of those disciplines can serve as a validation of the models used, especially in cases where analysis of the mathematical models reaches a point of infeasible difficulty \cite{holcombe2006agent}.

In this paper, we demonstrate a methodology by which the content validity of pedagogical tools, specifically interactive modeling environments in domains that deal heavily with modeling, curves and simulation, can be justified.
As an example, we apply this methodology to VERA due to its function as a tool that can produce simulation output time series that match common population curves that students learn about in biology ecology classes.
By using VERA, students often create conceptual models and simulate them in order to get a time series that aligns with what would be the population graph of a mathematical model.
In short, we will demonstrate that the conceptual model and simulation created by VERA will also reaffirm students' understanding of mathematical ecological models.
To do this, we implement two approaches, Hierarchical Clustering and Curve Fitting.
Hierarchical clustering functions as a bottom-up unsupervised method meaning that natural patterns in the data will emerge\cite{murtagh2012hierarchical, kassambara2017hierarchical}.
Curve Fitting was chosen because it is a supervised, top-down approach to classifying the time series.
This method means that we can take inspiration from known curves used in ecological modeling.
By finding a series of patterns in the hierarchical clustering that match, to a necessary threshold of accuracy, a group of chosen curves used in ecological modeling we can affirm that the simulation output of VERA would be useful for reinforcing population curves that a student learns in the classroom.

The VERA output of population graphs in the form of time series data can be difficult to interpret due to the presence of noise, feature correlation, and high dimensionality due to the length of the time series \cite{kotsakos2013time, aigner2011visualization}.
As mentioned earlier, time series analysis introduces unique challenges compared to other forms of data. 
One of the more significant factors is that VERA models and simulates complex ecological systems, so therefore outcomes may be vastly different and  there may be potential noise among relatively consistent outcomes \cite{anand2010complex}.
Another important consideration is that, in general, data samples may consist of different lengths of time, and different levels of starting populations for the species.
When combined, these and other factors can result in great difficulty adapting to the variability of samples that should belong to the same class \cite{bravi2011timeseries, Lhermitte2011timeseries}. 

To address these issues, we develop and compare the performance of two classification methodologies: (1) hierarchical clustering with dynamic time warping and (2) curve fitting, to identify patterns in species population outputs.
We first explore the use of hierarchical clustering with dynamic time warping to extract features in order to identify patterns in the population outputs \cite{luczak2016hierarchical}.
Hierarchical clustering seeks to identify groups of similar data samples in unlabeled data using a distance/similarity measure \cite{murtagh2012hierarchical}.
Hierarchical clustering can be agglomerative, meaning it starts with every sample in their own cluster and merges them until they form one cluster or divisive, which does the opposite \cite{omran2007overview}. 
{\L}uczak explored using hierarchical approaches on a variety of time series data sets \cite{luczak2016hierarchical}. 

Another popular approach that can be used to classify time series data is curve fitting, which consists of applying mathematical functions to fit the data to a particular curve.
These can be easier to adapt than empirical methods and can effectively suppress noise \cite{zeng2020review}. 
However, they can still struggle to adapt to data that does not align with standard function curves \cite{zeng2020review}. 
Curve fitting has been used extensively for extracting vegetation phenological metrics with different curve types including Logistic \cite{cao2015improved} and Gaussian \cite{jonsson2002seasonality} curves.
The curve-fitting approach took inspiration from common curves used to describe population dynamics found in ecological modeling.
As part of our initial experiments, see Section \ref{sec3}, we considered six patterns, exponential growth, capped growth, oscillation, dying off, Gaussian and constant.
A wide variety of possible curves were tested, however, the set of curves with the greatest success rate were those that match the patterns described above.

Across our tested sample, the curve fitting labeling and the hierarchical clustering with a second KNN corroboration step converged on the same curve type 89.38 percent of the time in the test set. 
For the rest of this paper, we will refer to this convergence between methods as accuracy.

\section{Background}
\subsection{VERA} \label{sec2:a}
VERA is an ecological modeling framework that allows users to create conceptual models and simulate their outcome \cite{an2018vera}. 
The conceptual models are expressed in the Component-Mechanism-Phenomenon (CMP) language, which is an extension of Structure-Behavior-Function models. \cite{joyner2014mila, goel2015impact, goel2009sbf}. 
This format of modeling allows for components such as biotics, abiotics, and habitats to be defined as well as relationships between them \cite{an2018vera}. 
Each component type has a set of adjustable parameters allowing for specialization to particular species. 
In order to assist in using valid parameters, users are able to lookup species with preset values from the Encyclopedia of Life (EOL)\cite{parr2014encyclopedia}. 
Once a model has been built, VERA translates the conceptual model into Netlogo \cite{tisue2004netlogo} primitives using a compiler. 
Running the simulation allows the user to see a plot see Fig. 2 of the population changes over time and they can then iterate through model-simulate-refine loops \cite{white1990causal} to refine their model.

VERA is used by a variety of stakeholders including students from Biology and Computer Science students at Georgia Institute of Technology (GATECH), Ecology Students at North Georgia Technical College (NGTC), and self directed learners that can find the site through the Smithsonian run EOL.
The variety in stakeholders corresponds to learners with different goals.
For example, the Computer Science students at Georgia Tech may use VERA to explore a scientific method of thinking as part of a cognitive science class, while an ecology student at North Georgia Technical College may model in VERA to explore the possibility space of an ecology.
Self-directed learners are different from both aforementioned groups in the facts that their goals remain unknown.
Nevertheless, the model-simulate-refine loops implicit in VERA's design encourage learning where feedback is necessary for understanding the conceptual modeling system, and therefore the effects of ones prospective modeled hypothesis.

\subsection{Related Works}

To the best of the authors' knowledge, we have not come across studies that categorize the time series of ecological simulation models in the context of education and its value in the use of pedagogical tools. 
This research lies at an understudied convergence of different topics.
As described by Cook \cite{cook2016validity}, while simulation based tools are becoming increasingly more popular, there is a lack of validation work proving these tools usefulness.
Since the publishing of the above source, there has been a number of studies that discuss the validity of simulation tools in educational contexts, but primarily in the medical field, and many of those leave validation studies for future work \cite{bogomolova2021validity, mcgrath2018validity, kong2021validity}.
However this does not mean that there is not value in doing validity studies for educational tools in other fields.

Borrowing from a validity framework developed by Messick used previously for psychological construct assessment, but cited in the studies above, there are six aspects of validity: content, substantive, structural, generalizability, external, and consequential \cite{messick1995validity}.
This research particularly focuses on content validity which is concerned with whether the assessment, or in this case the interactive modeling environment, resembles the domain in questions.
As shown by Cook in an earlier paper concerning what counts as validity evidence, the two most popular forms of content evidence were "group consensus or expert review", and creating an instrument "based on (or modified from) a previously validated instrument" \cite{cook2014validty}.
This research only concerns the second, however the first is mentioned briefly in discussion.
The previously validated instrument in this case is the domain of ecological modeling, and by reproducing validity tested methods in VERA we hope to extend content validity from the methods to the software itself.

Approaching the problem from the opposite direction, Holcombe suggests that "formally defined agents [within agent based systems] could be a basis for the validation and verification of models in the longer term" \cite{holcombe2006agent}.
Framed around the VERA software, this would take the form of ecological modeling validating work in their field using a tool such as VERA.
There are a number of studies in environmental studies and ecology that take directly this approach with VERA's underlying application NetLogo \cite{jaxarozen2019netlogo, thiele2012netlogo}.
Other existing research with this focus exists in the fields of economics, and social sciences \cite{windrum2007validation,ormerod2006validation}.
However, a literature review did not reveal any research with a focus on validation for pedagogical purposes in non-medical domains.
Rather than attempting to validate known models through agent based simulations, this
research seeks to use the convergence of both analytics tools as a method to determine pedagogical content validity in interactive learning environments.

\subsection{Ecological Population Behaviors}
Curves are used to describe population behaviors in the field of ecological modeling. 
The most foundational of these curves is the exponential growth curve which describes a closed population, modeled continuously with a positive instantaneous rate of increase \cite{Gotelli1995exp}. 
Exponential growth can typically be thought of as a new species being introduced to an environment, like an invasive species. This species may have ample food supplies and no natural predators initially, leading to rapid population growth \cite{snider2013introduction}.
Certain types of living organisms can also be geared more towards exponential growth such as bacteria because of their quick reproductive cycle.

Conversely, a negative instantaneous rate of increase creates the exponential decay curve.
This occurs when a species is exposed to an environment that does not support it, the population can rapidly decline \cite{winterhalder1997popdecline}.

However, the exponential curves assume infinite resources for consumption, unlimited reproduction, and density independent birth and death rates. 
Limiting those assumptions by capping resources and modifying the birth and death rates to be density dependent creates the foundation for the logistic curve \cite{Gotelli1995log}.
Another curve which is used to describe the slowing growth rate of populations over time is the sigmoid curve.
In ecological modeling, there is discussion of whether the the sigmoid curve would be a better representation of the circumstances used to describe the logistic curve above \cite{gamito1998sigmoid}.
The sigmoid curve is also described by an exponential curve which slows in growth as it reaches a population cap.
For the purposes of this paper, when describing curves, we will refer to both the logarithmic curve and the sigmoid curve as capped growth as that is an accurate descriptor for the circumstance which creates both curves.

Another model that includes species density and non closed populations in the form of environmental gradients is the Gaussian curve \cite{gaugh1974gaussian}.
A Gaussian curve could represent a species moving into a new environment where resources are plentiful and then consuming at an unsustainable rate such that the population then falls to a more sustainable level.

Populations in nature are also subject to temporality in that there may be resource abundant seasons or migration and location dependent mating.
In the case of resource abundant seasons, the consumer can be said to be CR coupled with their resource \cite{vandermeer2004oscillation}.
This creates our oscillation curve.
In the case of VERA, our oscillation curve is caused by the interactions with predator populations which becomes cyclical when the random interaction between populations reaches a stable equilibrium such that neither population falls below a critical minimum population size.
These behaviors explain five of the six patterns we identify as having relevant ecological meaning as these five curves deal primarily with the biotic component of VERA.
We are also proposing another curve, constant, for use in classification as it accounts for the abiotic component in VERA. \cite{netLogoUserManual}.
This could be average number of square inches of rain per year or vital non-biotic nutrients in the soil.
Lastly we will be using a category for all population graphs which do not fit into the above curves for which we will refer to as outliers.

\section{Methodology}  \label{sec3}
\subsection{Data processing}
The data set was manually downloaded from VERA, first by identifying a user, cloning each of their models and then running a simulation to create the data.
The time series created by the simulation was then downloaded as a CSV.

In order to account for variations in the population scales in the graphs, which can go from 0-25,000, we scale the data from zero to one based on the max population value in the graph. 
All data lengths are also cut to 400 values (400 months in simulation), so that the samples are all the same length. 
This is the default simulation length for VERA, however students are able to lengthen or shorten the default simulation length.
Graphs shorter than 400 data points were not included in this sample.

Three different user sets were looked at in this analysis, and they were chosen primarily based on the reasoning that their differing motivation for using VERA would lead to differing types of modeling behavior.
The first user set was Computer Science students from Georgia Tech (GATECH). As part of the cognitive science course, students are asked to use VERA and describe how it can assist with "scientific thinking".
Twenty students were then randomly selected from the course rosters of the Spring and Summer 2022 sessions for the course.
Every single model from a chosen student was then used in the data set resulting in 177 models and 685 species/curves total.

The second user set was students studying ecology at North Georgia Technical College (NGTC). 
Every student who took an ecology course that used VERA across the Summer and Fall 2022 was included in the data set.
In total this was 19 students, creating 57 models, including 194 species.

This last user group was Self-Directed Learners (SDLs). 
This is our classification for learners who are not tied to any known classroom use of VERA.
In order to determine this, all users of VERA were pulled, and any that contained a GATECH email, a NGTC email, or had email that had a first and last name that matched a known student were removed from the data.
This was evaluated by any new users from January 1st of 2022 to September 1st 2023.
The resulted in 55 users, of which only 23 had actually created a model.
From those 23 users, 29 models were pulled, containing 92 species.

A superset of all of the above user groups was also evaluated consisting of 263 models and 971 total species. Below is a table showing the curves broken up by type and user group.

\begin{table}[h]
\centering
\def\arraystretch{1}
\begin{tabular}{|c|c|c|c|c|c|c|c|c|}
\hline
& \multicolumn{2}{|c|}{\textbf{GATECH}} & \multicolumn{2}{|c|}{\textbf{NGTC}} & \multicolumn{2}{|c|}{\textbf{SDLs}} &                        \multicolumn{2}{|c|}{\textbf{All}} \\ \cline{2-9}
\multirow{-2}{*}{\textbf{Curve}} & \multicolumn{1}{|c|}{\#} & \% & \multicolumn{1}{|c|}{\#} & \%  & \multicolumn{1}{|c|}{\#} & \% & \# & \% \\ \hline
Exponential & 17 & 2.48 & 11 & 5.67 & 1 & 1.08 & 29 & 2.99 \\ \hline
Capped Growth & 48 & 7.01 & 18 & 9.28 & 9 & 9.78 & 75 & 7.72 \\ \hline
Exponential Dying & 287 & 41.89 & 92 & 47.42 & 41 & 44.56 & 420 & 43.33 \\ \hline
Oscillation & 28 & 4.09 & 28 & 14.43 & 13 & 14.13 & 69 & 7.10 \\ \hline
Constant & 145 & 21.17 & 2 & 1.03 & 15 & 16.30 & 162 & 16.68 \\ \hline
Gaussian & 127 & 18.54 & 33 & 17.01 & 9 & 9.78 & 169 & 17.40 \\ \hline
Outlier & 33 & 4.82 & 11 & 5.67 & 4 & 4.34 & 47 & 4.84 \\ \hline
\textbf{Total} & 685 & 100 & 194 & 100 & 92 & 100 & 971 & 100 \\ \hline
\end{tabular}
 \caption{Graphs broken up by curve type and data set, number \# and percent \%}
 \label{curve-sets}
\end{table}

\subsection{Old Training Set}
The lab had previously done research into self directed learners in order to better understand the types of models created by people outside a classroom setting where we know the models that they will create due to the assignment.
Overall, it was discovered that many users copied the exemplar models, or created models exhibiting the same basic structure but used their own species . \cite{an2023dissertation}
To understand the behaviors of the models, we wanted to align the simulation behaviors with what is expected for certain ecological relationships.
For example, oscillation may represent a predator-prey relationship where the populations cycle with each other.
While these types of relationships can be easily defined in an ecological sense, it was not clear how to align them with the simulation outputs.
In order to do this, we explored both top down and bottom up clustering of the simulation outputs to search for meaningful patterns. 
While searching, we utilized the silhouette score from KNN to identify meaningful clusters. 
After some manual work comparing the strongest clusters to our ecological behaviors, we identified six patterns present in the different clusters. 
These are the six categories we use: constant, dying off, oscillation, exponential growth, and capped growth, and gaussian.
The curve fitting approach was then developed with these meaningful clusters in mind, however many different other types of curves were tested but were found to be unsuccessful.
In the figure below, we can see all the different curve types, not including constant which would just be a straight line across the top.
While this classification list is non-exhaustive of all the curves in ecological modeling, we will see that it is comprehensive enough for testing purposes.

\begin{figure}[h]
    \centering
    \includegraphics[width=.8\linewidth]{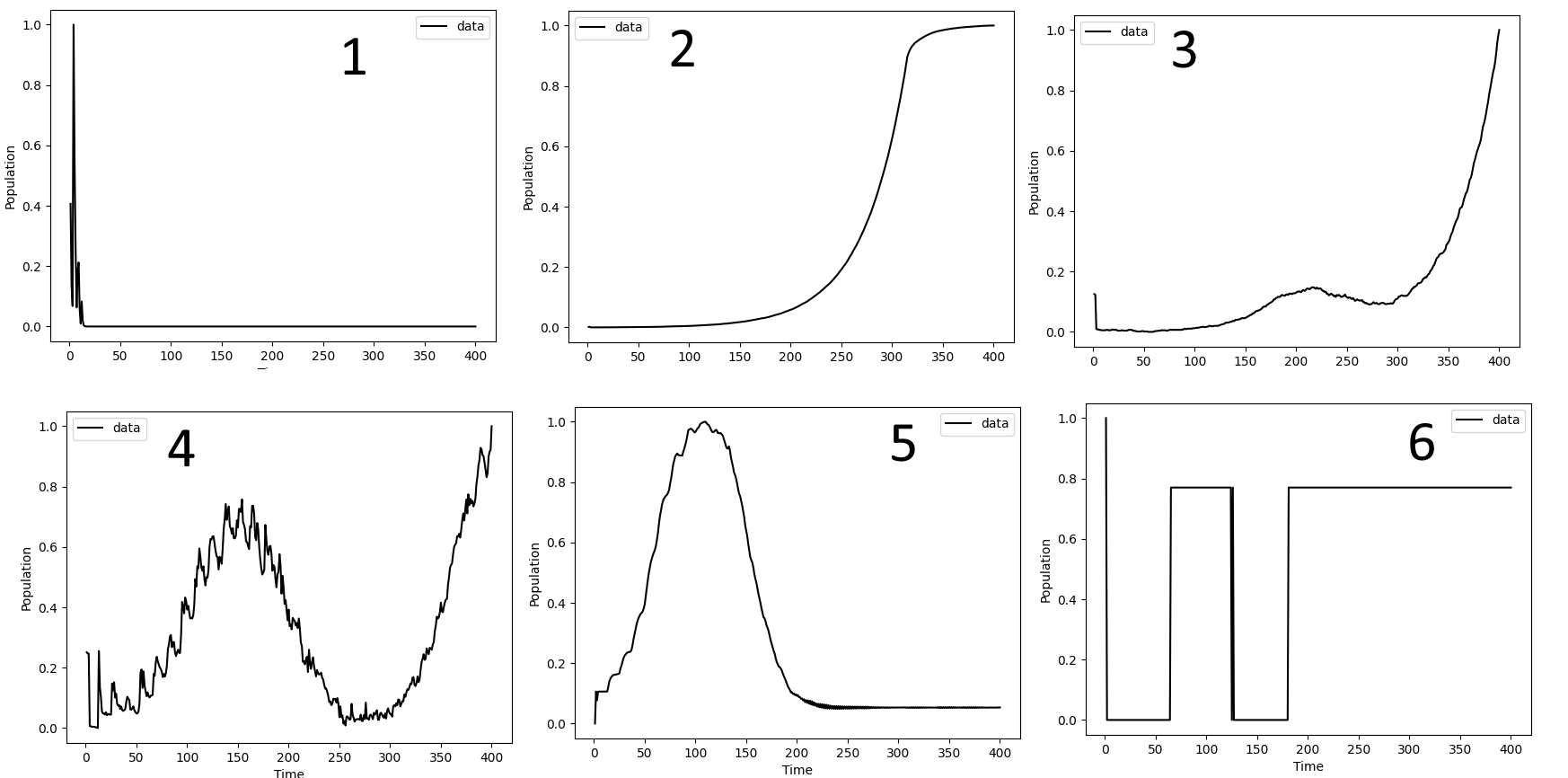}
    \caption{Curve types as labeled (1) dying, (2) capped growth, (3) exponential, (4) oscillation, (5) gaussian, (6) outlier}
    \label{fig:curve-types}
\end{figure}

The old training set was comprised of self directed learners from January 1st of 2019 to December 31st of 2021. It was made up of 197 models, which comprised 724 species.
We ran the two methods on this data set originally and use it to set the hyperparameters for the system.

\subsection{Hierarchical clustering}
In order to cluster the data, we use agglomerative hierarchical clustering with complete linkage, meaning clusters are merged based on the farthest distance between points in the clusters.
Similar to work by An, we use dynamic time warping with euclidean distance in order to calculate a distance matrix on the training data. \cite{an2023dissertation}
To isolate the patterns that are subject variation, we remove all of the constant curves from the clustering.
This is done by creating a set in python with all the values from the curve.
If the set length is one, then we know that the values of the curve do not change and therefore it is a constant curve.

Additionally, the data set is broken into a training set made up of 70\% of the data determined by hierarchical clustering, and a test set made up of the remaining 30\% determined by KNN on the clusters. 
The purpose of this is two fold.
The first is to evaluate the quality of the hierarchical clusters.
If we introduce new curves and run KNN to determine their similarity to existing clusters, then the introduced curve should fall into a similar cluster to that of their curve type.
Second, by evaluating an introduced curve by KNN of the medoid curve, a representative curve which has the least distance between itself and the other curves, in each cluster, we can speed up the process of comparison, allowing for quicker classification in the future.
The KNN process is described further in the section below.

With the remaining curves, a dendrogram is created using the scipy dendrogram function and then it is flattened using the scipy fcluster function.
This separates the dendorgram into clusters based on euclidean distance.
Within each cluster a prototypical representative, the mediod, is found determining which curve has the least distance between itself and all of the other members.
For classification, clusters are categorized based on the label attached to the representative curve based on the label from the curve fitting algorithm.
The label for each member of the cluster is then also evaluated in order to see if it matched the representatives' label in more than 55\% of cases.
If the 55\% threshold was not met, the entire cluster was classified as outlier.

\begin{figure}
    \centering
    \includegraphics[width=.7\linewidth]{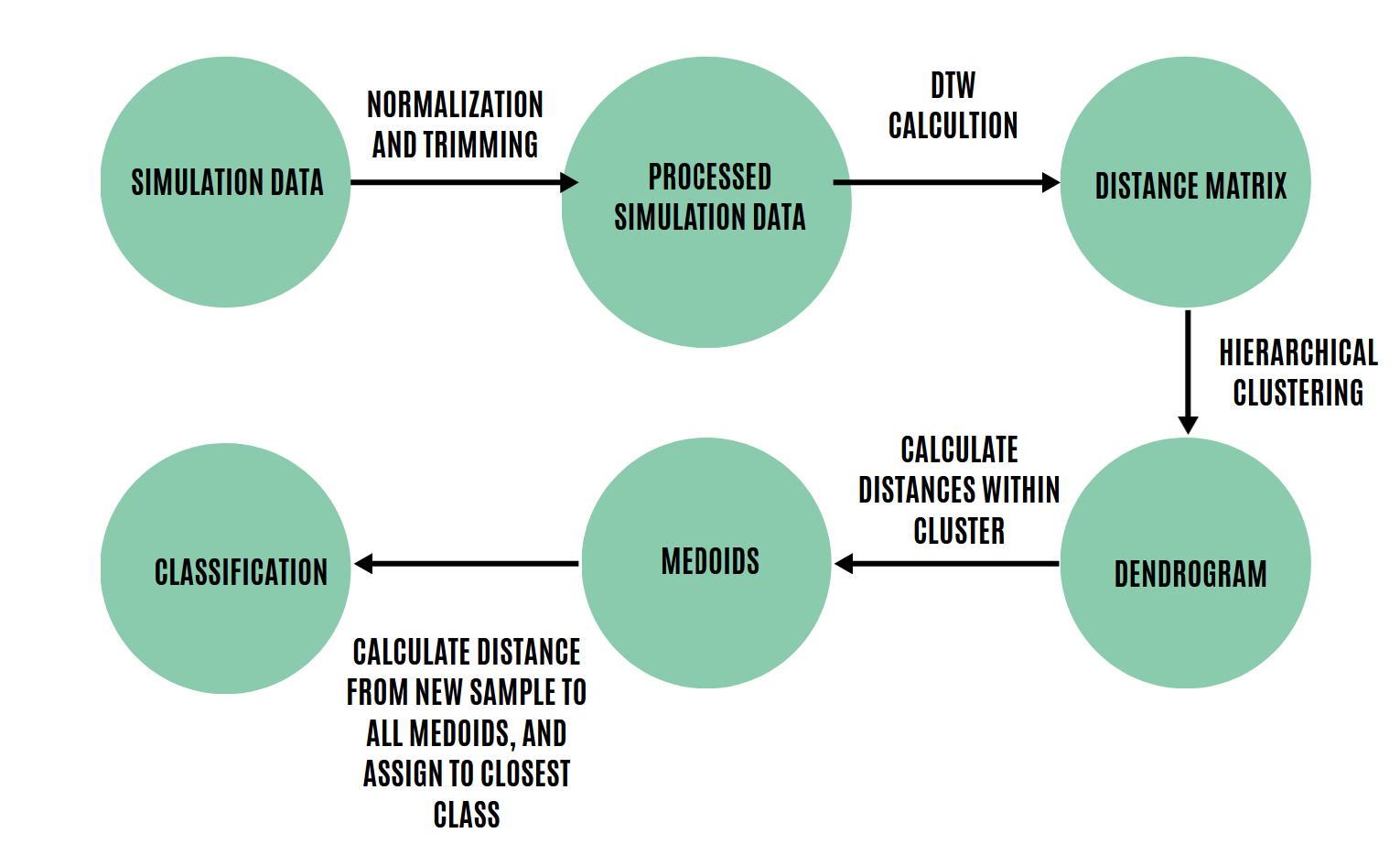}
    \caption{Hierarchical Clustering Process}
    \label{fig:clustering_process}
\end{figure}

\subsubsection{KNN Classification}

Since clustering is an unsupervised problem, we needed to develop another approach in order to perform classification on new samples using the clusters. 
For this, we first chose to pick medoid samples from each cluster, this allows us to speed up the computation time by only comparing against a subset of the data, since we desire to use this analysis for live feedback in VERA. 
Medoids are chosen as the data point with the minimum average distance to all other samples in a cluster.
Given a new data sample, we calculate the distance from this sample to all of the medoids and assign the class as the same as the medoid it has a minimum distance to. 
In this sense, the classification can be thought of as performing KNN on the medoids where k=1. 
There is also a minimum distance threshold that identifies a sample as an outlier if it is not close enough to any of the medoids.
This minimum distance threshold value was set to 5.

\subsection{Curve Fitting}
Curve fitting was undertaken by using the expected fundamental curves and inserting their mathematical equation, and the time series into Scipy's curve fit function. 
Curve fit then tries to return parameters for the mathematical equation that minimizes the residual sum of squares such that it fits to the given time series.
Each time series was then tested against every curve type in order to detect which curve type minimizes error. 

\begin{figure}[h]
    \centering
    \includegraphics[width=.7\linewidth]{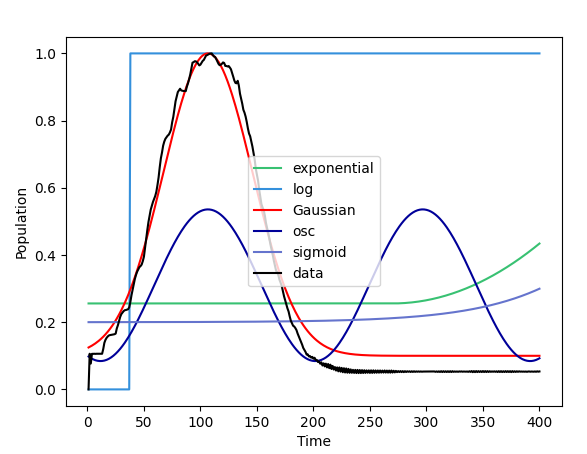}
    \caption{Visualization of Curve Fitting}
    \label{fig:curve-fit}
\end{figure}

Due to the variability in scale of the population graphs and the multidimensionality of the parameter space, four methods were taken to ensure that the function was successful in finding matching curves. 
First, the population graphs were normalized to the maximum value of the graph. 
This is valuable because the curve fit function searches in depth for any parameter values between integers, and this shrinks the possibility space.
Limiting the search such that the range is zero to one instead of zero to twenty five thousand, which is around the largest maximum value in the VERA system, means that when curve fit searches into multiple decimal places for a given parameter, it is more likely to search successfully.
Second was offering a prototypical set of variables for each curve. This limited the bounds such that the higher probability for success parameter space was searched first. 
The prototypical set of variables, for example, meant that sin waves with a period exponentially longer than the time series length were not searched. 
Third the parameter space was divided and the function was executed for each subsection of the parameter space. 
This emphasized the divided parameters allowing for a more fine grain search as more search cycles were spent in a smaller area.
Lastly, for curves where simple rules properly outline their behavior, a rule based system was used.
In practice, this means that the constant and exponential dying curves were evaluated using rules. 
Constant was evaluated by checking if all values in the time series were the same and not zero, and exponential dying was evaluated by checking if the population reached zero at any point and remained there, or close, for the remainder of the time series.
Closeness to zero was determined by the hyperparameter of 4\% of the maximum population which was determined during the old training set.

In order to detect outliers, error was evaluated by measuring the residual sum of square between the fit curve and the given population graph. 
A hyperparameter of an error threshold was then manually tuned on the old training set in order find best results, which gave a value of 5.7.
Any curve above that threshold was then classified as a outlier.

\section{Results}
Preliminary findings in the form of the silhouette scores are discussed briefly followed by the outcomes of the this research.
Our major results includes a description of the results as well as an analysis of each of the user sets and superset. 
The analysis will consist of the dendrograms for each set as well as a description for both of the classification steps.
\begin{figure}[h]
    \centering
    \includegraphics[width=.7\linewidth]{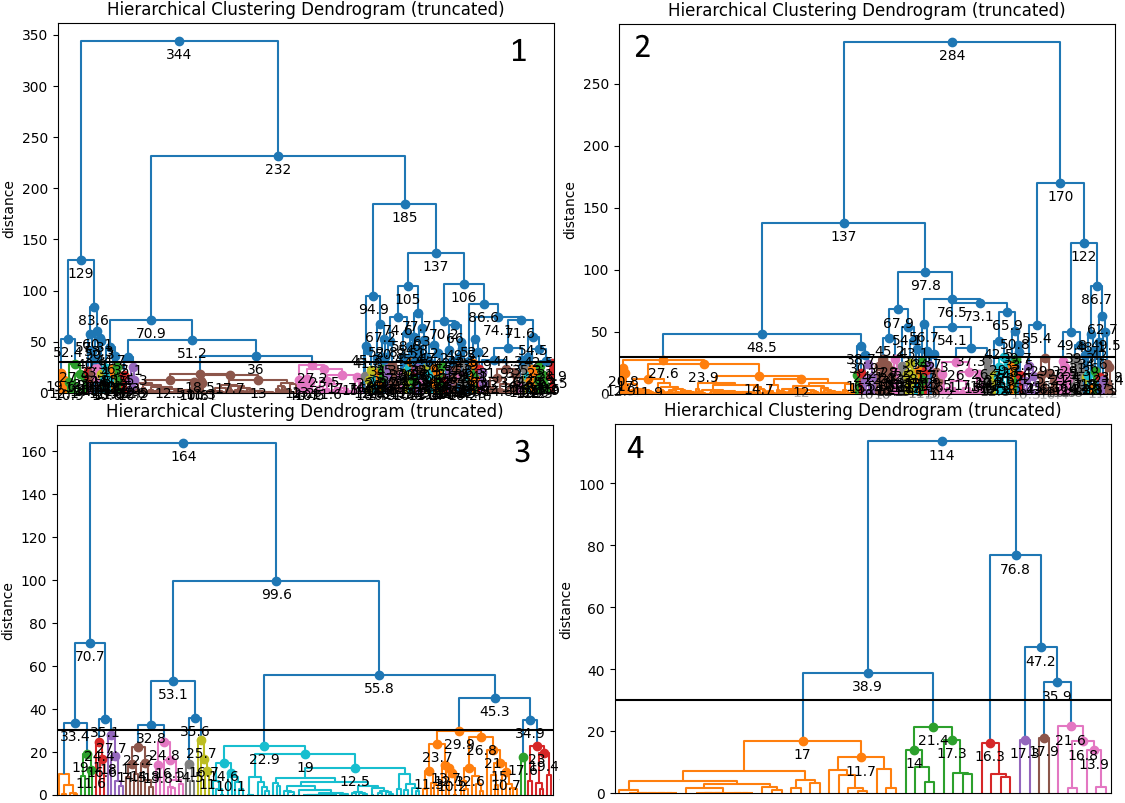}
    \caption{Dendrogram for all datasets. As numbered (1) superset, (2) GATECH, (3) NGTC, (4) SDLs.}
    \label{fig:dendrograms-all}
\end{figure}
\subsection{Silhouette Plot}
Below are the silhouette scores for Kmeans clusters on the test dataset. 
This helped us decide on our different curve types. As seen in the table below 7 clusters is the max number before a large drop off in score. 
This score then recovers in higher N values, but meaningful ecological differences between the clusters were not differentiable. 
These represent our six curves plus outliers.

\begin{table}
    \centering
    \def\arraystretch{1}
    \begin{tabular}{|c|c|c|c|c|c|c|c|c|c|} \hline 
         \textbf{Clusters} & 2 & 3 & 4 & 5 & 6 & 7 & 8 & 9 & 10 \\ \hline 
         \textbf{Silhouette Score} & 0.7292 & 0.6319 & 0.6502 & 0.6363 & 0.6478 & 0.6471 & 0.6191 & 0.6436 & 0.6444 \\ \hline
    \end{tabular}
    \caption{Silhouette scores based on cluster number (higher better)}
    \label{tab:silhouette-scores}
\end{table}

In the figures below we can see a representation of our clusters. 
For the seven clusters as seen on the left, we really only have one highly negative cluster, cluster 5. 
This represents our outliers which are not easily classified. 
As we move up to 8 clusters, we develop another highly negative cluster in cluster 7 which does not add any information value.
\begin{figure}[h]
    \centering
    \includegraphics[width=.8\linewidth]{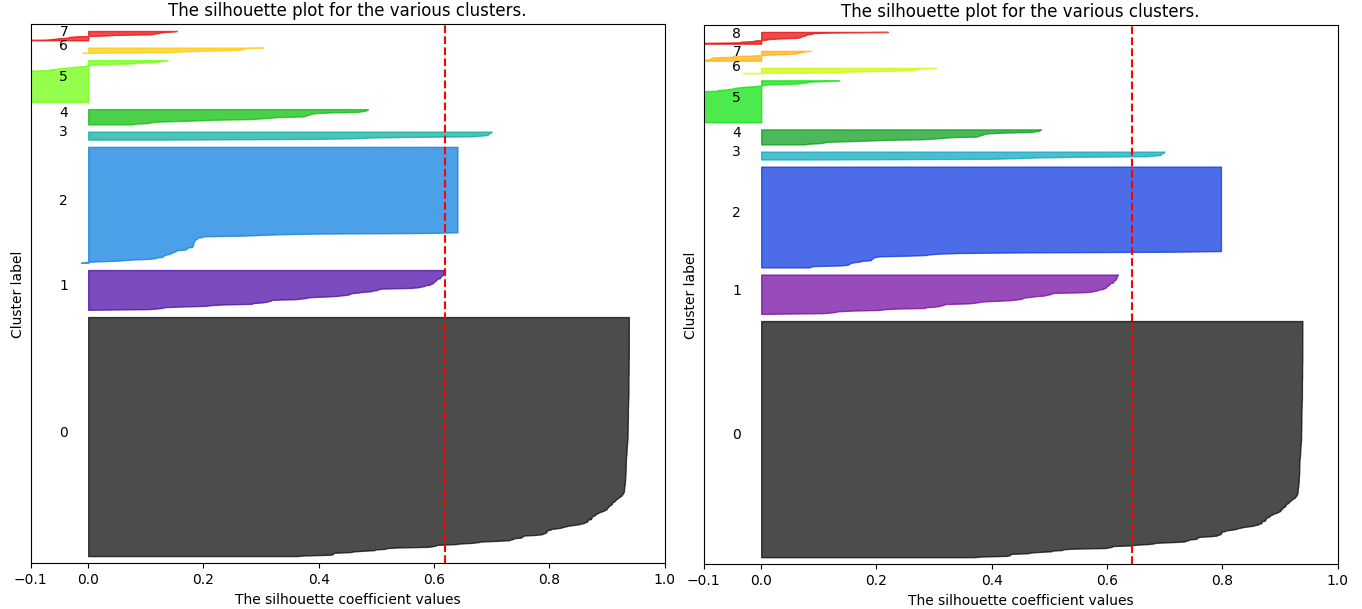}
    \caption{A comparison between 7 and 8 clusters}
    \label{fig:7-8-clusters}
\end{figure}
\subsection{Accuracy Findings}

\begin{table}[h]
    \centering
    \def\arraystretch{1}
    \begin{tabular}{|c|c|c|c|l|} \hline 
         \textbf{User Set} &  GATECH &  NGTC & SDLs & ALL \\ \hline 
         \textbf{Method Accuracy (\%)} &  88.43 &  71.65 &  70.65 & 89.38 \\ \hline
    \end{tabular}
    \caption{Table showing user sets and the percentage of agreement between the two methods on test set}
    \label{tab:resutls_table}
\end{table}

As shown in Table 3, the accuracy decreases when clustering is applied to the smaller sets of data. This is somewhat expected as more data would make for more numerous clusters as well as more defined clusters.

Using  30 as the distance measure for which the clusters are evaluated, as decided during the old training set, a total of 66 clusters were created.
Of these clusters 9 were labeled capped growth, with 59 of 61 of the curves being labeled as capped growth. 
4 of the clusters were exponential growth, with 17 of the 17 curves in the clusters being labeled as such. 
11 of the clusters were labeled exponential dying with 309 of the 320 curves matching. 
Oscillation made up 11 clusters with 57 of the 71 curves matching. 
Gaussian made up 22 of the clusters with 124 of the 136 curves matching. 
Outliers made up 9 of the clusters with 28 of the 32 matching.
This method of classification had a success rate of 93.24\%

10 additional of the above clusters were classified as outliers due to not meeting the 55\% matching threshold per cluster. 
These were clusters 3, 17, 21, 27, 30, 33, 40, 53, 54, 65.
Of these, 7 were originally Gaussian clusters, 1 was a dying cluster and 1 was a Oscillation cluster.
For the KNN validation test set, the accuracy fell to 89.38\%. 
This is followed by a confusion matrix, where the rows are the expected label and the columns are the final classification to the right of the total, and then the number of correct and incorrect to the left of the total.

\begin{table}[h]
\centering
\def\arraystretch{1}
\begin{tabular}{|c|c|c|c|c|c|c|c|c|c|c|}
\hline
\textbf{Label / Class} & \textbf{Out} & \textbf{Exp} & \textbf{Cap} & \textbf{Die} & \textbf{Osc} & \textbf{Gau} & \textbf{Con} & \textbf{Total} & \textbf{\# Correct} & \textbf{\# Incorrect} \\ \hline
Outlier                                &  3&   2&   6&  1&  0&   0  &  0     & 12& 3&9\\ \hline
Exponential                            &  0   &  8&  0   &   1&  0   &   1&  0    & 10& 8&1\\ \hline
Capped Growth                          &   0  &  0   &   22&  0   &   0  &  0   &  1& 23& 22&1\\ \hline
Dying                                  &   0  &  0   &  0   &  117&   0  &  3& 0     & 120& 117&3\\ \hline
Oscillation                            &  0   &  0   &  0   &   0  &  15&  4& 0     & 19& 15&4\\ \hline
Gaussian                               &   0&  1&  1&   2&  4&  42&  0    & 50& 42&8\\ \hline
Constant                               &  0   &  0   &  0   &  0   &  0   &   0  &  54& 54& 54&0\\ \hline
\end{tabular}
\caption{Confusion matrix for test set of superset, with number of correct and incorrect attached. }
\label{tab:superset-confusion}
\end{table}

Analyzing the Georgia Tech (GATECH) data set, a total of 35 clusters were created.
Of these clusters, 5 were labeled capped growth, with 39 of 48 of the curves being labeled as capped growth. 
3 of the clusters were exponential growth, with 13 of the 22 curves in the clusters being labeled as such. 
8 of the clusters were labeled exponential dying with 209 of the 216 curves matching. 
Oscillation made up 2 clusters with 8 of the 14 curves matching. 
Gaussian made up 13 of the clusters with 86 of the 97 curves matching. 
Outliers made up 5 of the clusters with 20 of the 23 matching.
This method of classification had a success rate of 90.57\%

7 additional of the above clusters were classified as outliers due to not meeting the matching threshold per cluster. 
These were clusters 4, 6, 7, 9, 20, 25, 27.
Of these, 3 were exponential, 1 was dying, 2 were Gaussian, and 1 was oscillation.
For the KNN validation test set, this fell to 88.46\%. 

\begin{table}[h]
\centering
\def\arraystretch{1}
\begin{tabular}{|c|c|c|c|c|c|c|c|c|c|c|}
\hline
\textbf{Label / Class} & \textbf{Out} & \textbf{Exp} & \textbf{Cap} & \textbf{Die} & \textbf{Osc} & \textbf{Gau} & \textbf{Con} & \textbf{Total} & \textbf{\# Correct} & \textbf{\# Incorrect} \\ \hline
Outlier                                &  2&   0  &   0&  0&  1&   1&  0& 4& 2&2\\ \hline
Exponential                            &  3&  0   &  1&   0  &  0   &   0&  0    & 4& 0&4\\ \hline
Capped Growth                          &   1  &  0   &   12&  0   &   0  &  0   &  0    & 13& 12&1\\ \hline
Dying                                  &   0  &  0   &  0   &  84&   0  &  1& 0     & 85& 84&1\\ \hline
Oscillation                            &  2&  0   &  0   &   0  &  1&  7& 0     & 10& 1&9\\ \hline
Gaussian                               &   2&  0   &  1   &   3&  3   &  40&  0    & 46& 40&6\\ \hline
Constant                               &  0   &  0   &  0   &  0   &  0   &   0  &  44& 44& 44&0\\ \hline
\end{tabular}
\caption{Confusion Matrix for test set of GATECH with number of correct and incorrect attached.}
\label{tab:gatech-confusion}
\end{table}

Analyzing the North Georgia Technical College (NGTC) data set, a total of 11 clusters were created.
Of these clusters four were labeled capped growth, with 19 of 22 of the curves being labeled as capped growth.
One of the clusters was exponential growth, with 6 of the 8 curves in the clusters being labeled as such. 
Two of the clusters were labeled exponential dying with 66 of the 85 curves matching. 
Oscillation made up one cluster with 8 of the 9 curves matching. 
Gaussian made up three of the clusters with 11 of the 16 curves matching. 
Outliers made up one of the clusters with 4 of the 5 matching.
This method of classification had a success rate of 78.62\%

1 additional of the above clusters was classified as an outlier due to not meeting the matching threshold. 
This cluster was cluster number 4, which was a Gaussian cluster.
For the KNN validation test set, this fell to 66.10\%. 
\begin{table}[h]
\centering
\def\arraystretch{1}
\begin{tabular}{|c|c|c|c|c|c|c|c|c|c|c|}
\hline
% Label / Class & Out & Exp & Cap & Die & Osc & Gau & Con  & Total& \# Correct&\# Incorrect\\ \hline
\textbf{Label / Class} & \textbf{Out} & \textbf{Exp} & \textbf{Cap} & \textbf{Die} & \textbf{Osc} & \textbf{Gau} & \textbf{Con} & \textbf{Total} & \textbf{\# Correct} & \textbf{\# Incorrect} \\ \hline
Outlier                                &  0&   0  &   0&  0   &  0   &   0  &  0     & 0& 0&0\\ \hline
Exponential                            &  0   &  2&  0   &   1&  0   &   0  &  0    & 3& 2&1\\ \hline
Capped Growth                          &   0  &  0   &   3&  0   &   0  &  0   &  0    & 3& 3&0\\ \hline
Dying                                  &   0  &  0   &  0   &  28&   0  &  0   & 0     & 28& 28&0\\ \hline
Oscillation                            &  5&  0&  0   &   0  &  3&  0   & 2& 10& 3&7\\ \hline
Gaussian                               &   3&  0&  0&   1&  1&  1&  7& 15& 1&14\\ \hline
Constant                               &  0   &  0   &  0   &  0   &  0   &   0  &  0& 0& 0&0\\ \hline
\end{tabular}
\caption{Confusion Matrix for the test set of NGTC with number of correct and incorrect attached.}
\label{tab:ngtc-confusion}
\end{table}

Analyzing the Self Directed Learners data set, a total of 5 clusters were created.
Of these clusters 2 were labeled capped growth, with 8 of 8 of the curves being labeled as capped growth.
0 of the clusters were labeled as capped growth.
2 of the clusters were labeled exponential dying with 31 of the 41 curves matching. 
Oscillation made up 1 cluster with 5 of the 7 curves matching. 
Gaussian made up 1 of the clusters with 2 of the 3 curves matching. 
Outliers made up 0 of the clusters.
This method of classification had a success rate of 81.35\%

2 additional of the above clusters were classified as an outlier due to not meeting the matching threshold. 
These were clusters 1 and 3, which were dying and Gaussian respectively.
After the KNN validation test set, this fell to 64.29\%. 
\begin{table}[h]
\centering
\def\arraystretch{1}
\begin{tabular}{|c|c|c|c|c|c|c|c|c|c|c|}
\hline
% Label / Class & Out & Exp & Cap & Die & Osc & Gau & Con  & Total & \# Correct&\# Incorrect\\ \hline
\textbf{Label / Class} & \textbf{Out} & \textbf{Exp} & \textbf{Cap} & \textbf{Die} & \textbf{Osc} & \textbf{Gau} & \textbf{Con} & \textbf{Total} & \textbf{\# Correct} & \textbf{\# Incorrect} \\ \hline
Outlier                                &   0&   0  &   1  &  0   &  0   &   0  &  0& 1& 0&1\\ \hline
Exponential                            &  0&  0   &  0   &   0&  0   &   0  &  0    & 0& 0&0\\ \hline
Capped Growth                          &   1&  0   &   2&  0   &   0  &  0   &  0    & 3& 2&1\\ \hline
Dying                                  &   0&  0   &  0   &  8&   0  &  0   & 2& 10& 8&2\\ \hline
Oscillation                            &  2&  0   &  0   &   0  &  4&  0   & 1& 7& 4&3\\ \hline
Gaussian                               &   2&  0   &  0&   0  &  0   &  0  &  1& 3& 0&3\\ \hline
Constant                               &  0   &  0   &  0   &  0   &  0   &   0  &  4& 4& 4&0\\ \hline
\end{tabular}
\caption{Confusion Matrix for test set of SDLs with number of correct and incorrect attached.}
\label{tab:sdl-confusion}
\end{table}

\section{Discussion and Future Work}

\subsection{Accuracy of different datasets}
As seen in the results section, the accuracy of the datasets is highly correlated with the size of the dataset. 
In smaller datasets, such as North Georgia Technical College and Self Directed Learners, the accuracy loss from the Clustering step to the KNN validation step was the largest, 12.52\% and 17.06\% respectively.
The small cluster size meant that many clusters in these datasets became infeasible during the KNN validation step.
This is seen in the Oscillation and Gaussian categories for the North Georgia Technical College dataset and the Exponential and Gaussian curves in the Self Directed Learners dataset.
This even occurred in the second largest data of Georgia Tech, where all the exponential curves were classified incorrectly.

In the largest dataset, we began to see a convergence of the clustering and KNN accuracies. 
The superset of data had an accuracy of 93.24\% during the clustering training step and then an accuracy of 89.38\% during the KNN validation step.
This is only a 3.86\% difference, which is significantly smaller than the smaller datasets.
This indicates that if we continue to increase our dataset size, we should see both values approach a similar accuracy percentage.

The largest challenge going forward, and a possible avenue for future work is the classification of outlier curves. 
The increased dataset size also meant that fewer clusters were classified as outliers during the step where 55\% of curves in a cluster had to share a label with the medoid curve or else it was classified as an outlier.
15.15\% of clusters were classified as outliers during this step in the superset, 20\% percent were classified as outliers for Georgia Tech, 9\% were classified as outliers in the for North Georgia Technical College (NGTC), and 40\% for self directed learners.
The lower percentage for NGTC outlier is unsurprising due to the level of guidance that was given while creating VERA models, their assignment is much more closed in scope.
Nevertheless, a larger dataset should converge on specific clusters for outlier curves of similar types, while simultaneously classify less clusters as outlier due to a lack of confluence in cluster labels.
Singular cases of outlier curves that do not follow a currently unknown outlier pattern will still be caught during the KNN step due to its likely distance from other curve types.

\subsection{Education Impacts}
This paper's analysis of VERA's simulation output is educationally valuable in three ways.
Firstly, the clustering demonstrates a mapping between VERA's simulation outputs and common ecological curves that some students may be learning in the classroom, validating VERA's use.
Secondly, and by extension, the output classes can be directly presented to learners to help them better understand the behavior of their models.
For users unfamiliar with common ecological curves, the output classes could give them a vocabulary stepping stone to better understand ecology.
Lastly, the largest motivational purpose of this work was in order to integrate the simulation output analysis into VERA's personalized coaching system. 
This would function by analyzing the user's model in comparison to other models in the database for which the simulation output is already known. 
The comparison would then allow us to suggest changes to the model which may assist the user in their learning process.
This could be useful in situations where a user is changing parameters or relationships in the conceptual model but still seeing the same class of simulation output, and therefore not gaining any new knowledge about the effects of their conceptual changes.
For example, when a user may have a model that consistently has species that are dying out, the personalized tutoring system may then suggest a node to be added or a parameter to be changed in order for the simulation to output an oscillation.
The change in simulation output, especially if the curves are labelled and viewable by the student, gives the student the ability to better conceptualize what changes in the ecology caused the change in the output.

\subsection{Validation}
For students who may be using VERA in biology or ecology classes, this research demonstrates an injective mapping of VERA's simulation output onto common curves found in mathematical ecological modeling.
While VERA is a platform primarily focused on conceptual modeling, making no claim to the accuracy of its simulations to real world ecology, it is still pedagogically useful as its outputs are likely to align with the mathematical models students are learning in their classrooms.
VERA is already partially justified in terms of content validity due to its use in ecology courses and its relation to the Smithsonian sponsored ecology website Encyclopedia of Life (EOL).
Despite that, this research developed a methodology by which modeling and simulation based pedagogical tools can be evaluated for content validity, and using VERA as an example, we were able to show successful overlap between these methods in larger datasets.

\subsection{Future Work}
The clearest direction for future work is the application of this research to a larger dataset.
This dataset will likely be the entirety of the VERA database, made of up over 7000 models.
A dataset of this size would be the best foundation for the model comparison part of VERA's personalized coaching system.

A proper analysis of outlier curves and their underlying causes in the VERA system is another path for future work. 
Patterns in these curves may reveal either more complex known ecological patterns currently unknown to the VERA team or it may reveal something corresponding to the software of VERA itself.
Further research into outlier patterns is the best way to differentiate between these two possible causes, and both lead to a greater understanding of the VERA system as a whole.

\section{Conclusion}
In this work, we explore analytics as a method by which we can assess content validity for conceptual modeling and simulation--based tools in domains that heavily rely on mathematical modeling.
As an example we apply the methodology to user generated content from VERA---an ecology education tool. 
Using common methods taken from both analysis of real world data, in the form of hierarchical clustering, and from mathematical ecological models, in the form of curve fitting, we analyze the outputs of an agent--based simulation system.
We found the greatest success using the largest data set reaching a 89.38\% success rate when compared to the curve fitting algorithm, with some indication that an even larger data set would perform better on average.
We also saw a convergence of the initial clustering algorithm success rate, and the KNN validation step, with the difference between the two evaluation methods approaching as little as 3.86\%.
The high rate of success in the data superset serves as a promising indication that this methodology could be applied to any other simulation--based education tools for which their domain has a strong modeling component, such as economics or other social sciences, and its application would give similar results for content valid systems.

\bibliographystyle{alpha}
\bibliography{main}

\newcommand{\etalchar}[1]{$^{#1}$}
\begin{thebibliography}{DlTOC{\etalchar{+}}21}

\bibitem[20222]{2022Modeling}
Modeling {Ecosystem} {Dynamics}, jun 8 2022.
\newblock [Online; accessed 2023-07-16].

\bibitem[ABH{\etalchar{+}}18]{an2018vera}
Sungeun An, Robert Bates, Jennifer Hammock, Spencer Rugaber, and Ashok Goel.
\newblock Vera: popularizing science through ai.
\newblock In {\em Artificial Intelligence in Education: 19th International Conference, AIED 2018, London, UK, June 27--30, 2018, Proceedings, Part II 19}, pages 31--35. Springer, 2018.

\bibitem[AGG{\etalchar{+}}10]{anand2010complex}
M~Anand, A~Gonzale, F~Guichard, J~Kolasa, and L~Parrott.
\newblock Ecological systems as complex systems: Challenges for an emerging science.
\newblock {\em Diversity}, 2:395--410, 2010.

\bibitem[AMST11]{aigner2011visualization}
Wolfgang Aigner, Silvia Miksch, Heidrun Schumann, and Christian Tominski.
\newblock {\em Visualization of time-oriented data}, volume~4.
\newblock Springer, 2011.

\bibitem[BA11]{bravi2011timeseries}
Seely~AJ. Bravi~A, Longtin~A.
\newblock Review and classification of variability analysis techniques with clinical applications.
\newblock {\em Biomedical engineering online}, pages 1--27, 2011.

\bibitem[BSM{\etalchar{+}}21]{bogomolova2021validity}
K.~Bogomolova, A.~H. Sam, A.~T. Misky, C.~M. Gupte, P.~H. Strutton, T.~J. Hurkxkens, and B.~P. Hierck.
\newblock Development of a virtual three‐dimensional assessment scenario for anatomical education.
\newblock {\em Anatomical sciences education}, 14, 2021.

\bibitem[CCST15]{cao2015improved}
Ruyin Cao, Jin Chen, Miaogen Shen, and Yanhong Tang.
\newblock An improved logistic method for detecting spring vegetation phenology in grasslands from modis evi time-series data.
\newblock {\em Agricultural and Forest Meteorology}, 200:9--20, 2015.

\bibitem[CH16]{cook2016validity}
D.~A. Cook and R.~Hatala.
\newblock Validation of educational assessments: a primer for simulation and beyond.
\newblock {\em Advances in simulation}, 1:1--12, 2016.

\bibitem[CZH{\etalchar{+}}14]{cook2014validty}
D.~A. Cook, B.~Zendejas, S.~J. Hamstra, R.~Hatala, and R.~Brydges.
\newblock What counts as validity evidence? examples and prevalence in a systematic review of simulation-based assessment.
\newblock {\em Advances in Health Sciences Education}, 19, 2014.

\bibitem[DB15]{dayblack2015simulation}
Crystal Day-Black.
\newblock Gamification: An innovative teaching-learning strategy for the digital nursing students in a community health nursing course.
\newblock {\em ABNF Journal}, 26, 2015.

\bibitem[DlTOC{\etalchar{+}}21]{delatorre2021simulation}
R.~De~la Torre, B.~S. Onggo, C.~G. Corlu, Nogal M., and A.~A. Juan.
\newblock The role of simulation and serious games in teaching concepts on circular economy and sustainable energy.
\newblock {\em Energies}, 14:1138, 2021.

\bibitem[ENB12]{evans2012predictive}
Matthew~R Evans, Ken~J Norris, and Tim~G Benton.
\newblock Predictive ecology: systems approaches, 2012.

\bibitem[Gam98]{gamito1998sigmoid}
S.~Gamito.
\newblock Growth models and their use in ecological modelling: an application to a fish population.
\newblock {\em Ecological modelling}, 113:83--94, 1998.

\bibitem[GJ15]{goel2015impact}
A~Goel and D~Joyner.
\newblock Impact of a creativity support tool on student learning about scientific discovery processes.
\newblock In {\em Proceedings of the Sixth International Conference on Computational Creativity}, 2015.

\bibitem[Got95a]{Gotelli1995exp}
N.~J. Gotelli.
\newblock {\em A primer of Ecology}.
\newblock Sinauer Associates Incorporate, 1995.

\bibitem[Got95b]{Gotelli1995log}
N.~J. Gotelli.
\newblock {\em A primer of Ecology}.
\newblock Sinauer Associates Incorporate, 1995.

\bibitem[GRV09]{goel2009sbf}
A.~K. Goel, S.~Rugaber, and S.~Vattam.
\newblock Structure, behavior, and function of complex systems: The structure, behavior, and function modeling language.
\newblock {\em Ai Edam}, 23, 2009.

\bibitem[HCR06]{holcombe2006agent}
M~Holcombe, S~Coakley, and Smallwood R.
\newblock A general framework for agent-based modelling of complex systems.
\newblock {\em Proceedings of the 2006 European conference on complex systems}, 1, 2006.

\bibitem[J.04]{vandermeer2004oscillation}
Vandermeer J.
\newblock Coupled oscillations in food webs: Balancing competition and mutualism in simple ecological models.
\newblock {\em The American Naturalist}, 163:857--867, 2004.

\bibitem[JE02]{jonsson2002seasonality}
Per Jonsson and Lars Eklundh.
\newblock Seasonality extraction by function fitting to time-series of satellite sensor data.
\newblock {\em IEEE transactions on Geoscience and Remote Sensing}, 40(8):1824--1832, 2002.

\bibitem[JGP14]{joyner2014mila}
David~A Joyner, Ashok~K Goel, and Nicolas~M Papin.
\newblock Mila--s: generation of agent-based simulations from conceptual models of complex systems.
\newblock In {\em Proceedings of the 19th international conference on intelligent user interfaces}, pages 289--298, 2014.

\bibitem[JHC74]{gaugh1974gaussian}
Gauch Jr, G.~Hugh, and Gene~B. Chase.
\newblock Fitting the gaussian curve to ecological data.
\newblock {\em Ecology}, 55, 1974.

\bibitem[JMA14]{janssen2014simulation}
\&~Waring T.~M. Janssen M.~A., Lee~A.
\newblock Experimental platforms for behavioral experiments on social-ecological systems.
\newblock {\em Ecology and Society}, 19, 2014.

\bibitem[JRKB19]{jaxarozen2019netlogo}
M.~Jaxa-Rozen, J.~H. Kwakkel, and M.~Bloemendal.
\newblock A coupled simulation architecture for agent-based/geohydrological modelling with netlogo and modflow.
\newblock {\em Environmental modelling \& software}, 115, 2019.

\bibitem[Kas17]{kassambara2017hierarchical}
A.~Kassambara.
\newblock {\em Practical guide to cluster analysis in R: Unsupervised machine learning}.
\newblock 2017.

\bibitem[KTGA13]{kotsakos2013time}
Dimitrios Kotsakos, Goce Trajcevski, Dimitrios Gunopulos, and Charu~C Aggarwal.
\newblock Time-series data clustering., 2013.

\bibitem[KW21]{kong2021validity}
S.~C. Kong and Y.~Q. Wang.
\newblock Item response analysis of computational thinking practices: Test characteristics and students’ learning abilities in visual programming contexts.
\newblock {\em Computers in Human Behavior}, 122, 2021.

\bibitem[LLE08]{eberhardt2008curves}
D.~P.~Demaster L.~L.~Eberhardt, J. M.~Breiwick.
\newblock Analyzing population growth curves.
\newblock {\em Oikos}, 117:1240--1246, 2008.

\bibitem[{\L}uc16]{luczak2016hierarchical}
Maciej {\L}uczak.
\newblock Hierarchical clustering of time series data with parametric derivative dynamic time warping.
\newblock {\em Expert Systems with Applications}, 62:116--130, 2016.

\bibitem[LVVC11]{Lhermitte2011timeseries}
S.~Lhermitte, J.~Verbesselt, W.W. Verstraeten, and P.~Coppin.
\newblock A comparison of time series similarity measures for classification and change detection of ecosystem dynamics.
\newblock {\em Remote sensing of environment}, 115:3129--3152, 2011.

\bibitem[MC17]{murtagh2012hierarchical}
F.~Murtagh and P.~Contreras.
\newblock Algorithms for hierarchical clustering: an overview.
\newblock {\em Wiley Interdisciplinary Reviews: Data Mining and Knowledge Discovery}, 2, 2017.

\bibitem[Mes95]{messick1995validity}
S.~Messick.
\newblock Validity of psychological assessment: Validation of inferences from persons' responses and performances as scientific inquiry into score meaning.
\newblock {\em American psychologist}, 50, 1995.

\bibitem[MTD{\etalchar{+}}18]{mcgrath2018validity}
J.~L. McGrath, J.~M. Taekman, P.~Dev, D.~R. Danforth, D.~Mohan, N~Kman, and K.~Won.
\newblock Using virtual reality simulation environments to assess competence for emergency medicine learners.
\newblock {\em Academic Emergency Medicine}, 25, 2018.

\bibitem[OES07]{omran2007overview}
Mahamed~GH Omran, Andries~P Engelbrecht, and Ayed Salman.
\newblock An overview of clustering methods.
\newblock {\em Intelligent Data Analysis}, 11(6):583--605, 2007.

\bibitem[OR06]{ormerod2006validation}
P.~Ormerod and B.~Rosewell.
\newblock Validation and verification of agent-based models in the social sciences.
\newblock {\em International workshop on epistemological aspects of computer simulation in the social sciences}, 2006.

\bibitem[oTDL23]{vera2023site}
Georgia~Institute of~Technology~Design and Intelligence Lab.
\newblock Virtual ecological research assistant (vera), 2023.

\bibitem[Ott03]{ottino2003complex}
Julio~M. Ottino.
\newblock Complex systems.
\newblock {\em American Institute of Chemical Engineers}, 49, 2003.

\bibitem[Pie81]{pielou1981stock}
E.C Pielou.
\newblock The usefulness of ecological models: a stock-taking.
\newblock {\em The Quarterly Review of Biology}, 56:17--31, 1981.

\bibitem[PWL{\etalchar{+}}14]{parr2014encyclopedia}
Cynthia~S Parr, Mr~Nathan Wilson, Mr~Patrick Leary, Katja~S Schulz, Ms~Kristen Lans, Ms~Lisa Walley, Jennifer~A Hammock, Mr~Anthony Goddard, Mr~Jeremy Rice, Mr~Marie Studer, et~al.
\newblock The encyclopedia of life v2: providing global access to knowledge about life on earth.
\newblock {\em Biodiversity data journal}, (2), 2014.

\bibitem[S.]{an2023dissertation}
An~S.
\newblock An, sung disseration.
\newblock Sung dissertation.

\bibitem[SB13]{snider2013introduction}
SB~Snider and JN~Brimlow.
\newblock An introduction to population growth.
\newblock {\em Nature Education Knowledge}, 4(4):3, 2013.

\bibitem[TKG12]{thiele2012netlogo}
J.~C. Thiele, W.~Kurth, and V.~Grimm.
\newblock Rnetlogo: An r package for running and exploring individual‐based models implemented in netlogo.
\newblock {\em Methods in Ecology and Evolution}, 3, 2012.

\bibitem[TW04a]{netLogoUserManual}
S.~Tisue and U~Wilensky.
\newblock Netlogo: A simple environment for modeling complexity.
\newblock {\em International conference on complex systems}, 21:16--21, 2004.

\bibitem[TW04b]{tisue2004netlogo}
Seth Tisue and Uri Wilensky.
\newblock Netlogo: A simple environment for modeling complexity.
\newblock In {\em International conference on complex systems}, volume~21, pages 16--21. Citeseer, 2004.

\bibitem[WB97]{winterhalder1997popdecline}
Lu~F. Winterhalder~B.
\newblock A forager‐resource population ecology model and implications for indigenous conservation.
\newblock {\em Conservation Biology}, 11:1354--1364, 1997.

\bibitem[WF90]{white1990causal}
Barbara~Y White and John~R Frederiksen.
\newblock Causal model progressions as a foundation for intelligent learning environments.
\newblock {\em Artificial intelligence}, 42(1):99--157, 1990.

\bibitem[WFM07]{windrum2007validation}
P.~Windrum, G.~Fagiolo, and A.~Moneta.
\newblock Empirical validation of agent-based models: Alternatives and prospects.
\newblock {\em Journal of Artificial Societies and Social Simulation}, 10, 2007.

\bibitem[ZWX{\etalchar{+}}20]{zeng2020review}
Linglin Zeng, Brian~D Wardlow, Daxiang Xiang, Shun Hu, and Deren Li.
\newblock A review of vegetation phenological metrics extraction using time-series, multispectral satellite data.
\newblock {\em Remote Sensing of Environment}, 237:111511, 2020.

\end{thebibliography}

\end{document}